\documentclass[letterpaper, 10 pt, journal, twoside]{IEEEtran}
\def\secref#1{Sec.~\ref{#1}}

\def\figref#1{Fig.~\ref{#1}}
\def\tabref#1{Tab.~\ref{#1}}
\def\eqref#1{Eq.~(\ref{#1})}

\usepackage[numbers,sort&compress]{natbib}
\usepackage{hyperref}

\usepackage{graphicx}
\usepackage{comment}
\usepackage{amsmath}
\usepackage{multirow}
\usepackage{tabularx}
\usepackage{pifont}
\usepackage{siunitx}
\usepackage{xcolor}
\definecolor{Red}{RGB}{192, 26, 38}
\definecolor{Orange}{RGB}{251, 151, 39}
\definecolor{Green}{RGB}{167,204,110}
\definecolor{Black}{RGB}{0,0,0}

\usepackage{amssymb}
\usepackage{xcolor}

\graphicspath{ {./figures/} }
\usepackage{xcolor}

\definecolor{MK_Two_One}{RGB}{178,24,43} % Nice red
\definecolor{MK_Two_Two}{RGB}{239,138,98}
\definecolor{MK_Two_Three}{RGB}{253,219,199}
\definecolor{MK_Two_Four}{RGB}{209,229,240}
\definecolor{MK_Two_Five}{RGB}{103,169,207}
\definecolor{MK_Two_Six}{RGB}{33,102,172} % Nice blue

\usepackage{hyperref}
\hypersetup{
colorlinks=true
,linkcolor=MK_Two_Six
,citecolor={green!80!black}
,filecolor=MK_Two_Six
,urlcolor= {blue!80!black}
,menucolor=MK_Two_Five
,runcolor=MK_Two_Four
,linkbordercolor=MK_Two_One
,citebordercolor=MK_Two_Two
,filebordercolor=MK_Two_Three
,urlbordercolor=MK_Two_Six
,menubordercolor=MK_Two_Five
,runbordercolor=MK_Two_Four
}

\usepackage{glossaries}
\usepackage[abbreviations]{glossaries-extra}
\usepackage[utf8]{inputenc}

\glssetcategoryattribute{abbreviation}{indexonlyfirst}{true}

\glssetcategoryattribute{abbreviation}{nohyperfirst}{true}

\newabbreviation{auroc}{AUROC}{Area Under the Receiver Operating Characteristic Curve}
\newabbreviation{accuracy}{Acc}{Accuracy}

\newabbreviation{cnn}{CNN}{Convolutional Neural Network}

\newabbreviation{fov}{FoV}{Field of View}
\newabbreviation{fpr}{FPR}{False Positive Ratio}

\newabbreviation{gnn}{GNN}{Graph Neural Network}
\newabbreviation{gcn}{GCN}{Graph Convolutional Network}
\newabbreviation{gru}{GRU}{Gated Recurrent Unit}
\newabbreviation{gmm}{GMM}{Gaussian Mixture Model}

\newabbreviation{imu}{IMU}{Inertial Measurement Unit}
\newabbreviation{irl}{IRL}{Inverse Reinforcement Learning}
\newabbreviation{iou}{IOU}{Intersection over Union}
\newabbreviation{knn}{KNN}{K-Nearest Neighbors}

\newabbreviation{lagr}{LAGR}{Learning Applied to Ground Vehicles}
\newabbreviation{lidar}{LiDAR}{Light Detection and Ranging}

\newabbreviation{mlp}{MLP}{Multi-Layer Perceptron}
\newabbreviation{mpc}{MPC}{Model Predictive Controller}
\newabbreviation{mse}{MSE}{Mean Squared Error}
\newabbreviation{mae}{MAE}{Mean Absolute Error}
\newabbreviation{ood}{OOD}{Out-Of-Distribution}
\newabbreviation{id}{ID}{In-Distribution}
\newabbreviation{rbf}{RBF}{Radial Basis Function}
\newabbreviation{rmp}{RMP}{Riemannian Motion Policies}
\newabbreviation{ros}{ROS}{Robot Operating System}
\newabbreviation{ros1}{ROS~1}{Robot Operating System}
\newabbreviation{roc}{ROC}{Receiver Operating Characteristic}
\newabbreviation{rf}{RF}{Random Forest}

\newabbreviation{sdf}{SDF}{Signed Distance Field}
\newabbreviation{slam}{SLAM}{Simultaneous Localization and Mapping}
\newabbreviation{svm}{SVM}{Support Vector Machine}
\newabbreviation{svc}{SVC}{Support Vector Classifier}
\newabbreviation{wvn}{WVN}{Wild Visual Navigation}

\newabbreviation{vit}{ViT}{Vision Transformer}

\begin{document}
%
% paper title
% Titles are generally capitalized except for words such as a, an, and, as,
% at, but, by, for, in, nor, of, on, or, the, to and up, which are usually
% not capitalized unless they are the first or last word of the title.
% Linebreaks \\ can be used within to get better formatting as desired.
% Do not put math or special symbols in the title.
\title{
Identifying Terrain Physical Parameters from Vision - Towards Physical-Parameter-Aware Locomotion and Navigation
}
%
%
% author names and IEEE memberships
% note positions of commas and nonbreaking spaces ( ~ ) LaTeX will not break
% a structure at a ~ so this keeps an author's name from being broken across
% two lines.
% use \thanks{} to gain access to the first footnote area
% a separate \thanks must be used for each paragraph as LaTeX2e's \thanks
% was not built to handle multiple paragraphs
%

% \author{Michael~Shell,~\IEEEmembership{Member,~IEEE,}
%         John~Doe,~\IEEEmembership{Fellow,~OSA,}
%         and~Jane~Doe,~\IEEEmembership{Life~Fellow,~IEEE}% <-this % stops a space
% \thanks{M. Shell was with the Department
% of Electrical and Computer Engineering, Georgia Institute of Technology, Atlanta,
% GA, 30332 USA e-mail: (see http://www.michaelshell.org/contact.html).}% <-this % stops a space
% \thanks{J. Doe and J. Doe are with Anonymous University.}% <-this % stops a space
% \thanks{Manuscript received April 19, 2005; revised August 26, 2015.}}
\author{Jiaqi Chen$^{1}$, Jonas Frey$^{1,2}$, Ruyi Zhou$^{1,3}$, Takahiro Miki$^{1}$, Georg Martius$^{2,4}$, and Marco Hutter$^{1}$% <-this % stops a space
% \thanks{*This work was not supported by any organization}% <-this % stops a space
\thanks{Manuscript received: March 30, 2024; Revised July 10, 2024; Accepted August 15, 2024.}
\thanks{This paper was recommended for publication by Editor A. Valada upon evaluation of the Associate Editor and Reviewers' comments. This project was supported by the Max Planck ETH Center for Learning Systems (Frey), Swiss National Science Foundation (SNSF) through project 166232, 188596, the European Union’s Horizon 2020 research and innovation program under grant agreement No.101016970 and No.101070405, and National Natural Science Foundation of China Youth Basic Research Project (Doctoral Student) under grant No.523B2039.}
\thanks{$^{1}$Jiaqi Chen, Jonas Frey, Ruyi Zhou, Takahiro Miki, and Marco Hutter are with Robotic Systems Laboratory, ETH Zurich, Zurich 8092, Switzerland (e-mail: chenjiaq@ethz.ch; jonfrey@ethz.ch; zhouryhit@gmail.com; tamiki@ethz.ch; mahutter@ethz.ch)}%
%\thanks{$^{2}$These authors are with the Robotic Systems Laboratory, ETH Zurich, Zurich
%8092, Switzerland (e-mail: jonfrey@ethz.ch; ruyzhou@ethz.ch; tamiki@ethz.ch; mahutter@ethz.ch)
%}
\thanks{$^{2}$Jonas Frey and Georg Martius are with Max Planck Institute for Intelligent Systems Tübingen, Tübingen 72076, Germany}
\thanks{$^{3}$Ruyi Zhou is with State Key Laboratory of Robotics and Systems, Harbin Institute of Technology, Harbin 150080, China}
\thanks{$^{4}$Georg Martius is with the University of Tübingen, Tübingen 72076, Germany (e-mail: georg.martius@tuebingen.mpg.de)}
\thanks{Digital Object Identifier (DOI): see top of this page.}
}

\markboth{IEEE Robotics and Automation Letters. Preprint Version. Accepted August, 2024}
{Chen \MakeLowercase{\textit{et al.}}: Identifying Terrain Physical Parameters from Vision} 

% The only time the second header will appear is for the odd numbered pages
% after the title page when using the twoside option.
% 
% *** Note that you probably will NOT want to include the author's ***
% *** name in the headers of peer review papers.                   ***
% You can use \ifCLASSOPTIONpeerreview for conditional compilation here if
% you desire.

% If you want to put a publisher's ID mark on the page you can do it like
% this:
%\IEEEpubid{0000--0000/00\$00.00~\copyright~2015 IEEE}
% Remember, if you use this you must call \IEEEpubidadjcol in the second
% column for its text to clear the IEEEpubid mark.

% use for special paper notices
%\IEEEspecialpapernotice{(Invited Paper)}

\bibliographystyle{unsrt}
% make the title area
\maketitle

% As a general rule, do not put math, special symbols or citations
% in the abstract or keywords.
\begin{abstract}
Identifying the physical properties of the surrounding environment is essential for robotic locomotion and navigation to deal with non-geometric hazards, such as slippery and deformable terrains.
It would be of great benefit for robots to anticipate these extreme physical properties before contact; however, estimating environmental physical parameters from vision is still an open challenge. 
Animals can achieve this by using their prior experience and knowledge of what they have seen and how it felt. 
In this work, we propose a cross-modal self-supervised learning framework for vision-based environmental physical parameter estimation, which paves the way for future physical-property-aware locomotion and navigation. We bridge the gap between existing policies trained in simulation and identification of physical terrain parameters from vision. We propose to train a physical decoder in simulation to predict friction and stiffness from multi-modal input. The trained network allows the labeling of real-world images with physical parameters in a self-supervised manner to further train a visual network during deployment, which can densely predict the friction and stiffness from image data. We validate our physical decoder in simulation and the real world using a quadruped ANYmal robot, outperforming an existing baseline method. We show that our visual network can predict the physical properties in indoor and outdoor experiments while allowing fast adaptation to new environments. 
--- Project Page \url{https://bit.ly/3Xo5AA8} ---
\end{abstract}

% Note that keywords are not normally used for peerreview papers.
% \begin{IEEEkeywords}
% IEEE, IEEEtran, journal, \LaTeX, paper, template.
% \end{IEEEkeywords}

% For peer review papers, you can put extra information on the cover
% page as needed:
% \ifCLASSOPTIONpeerreview
% \begin{center} \bfseries EDICS Category: 3-BBND \end{center}
% \fi
%
% For peerreview papers, this IEEEtran command inserts a page break and
% creates the second title. It will be ignored for other modes.
\IEEEpeerreviewmaketitle

\section{INTRODUCTION}
\label{sec:introduction}

\IEEEPARstart{L}{egged} robots excel in walking on challenging terrains~\cite{Lee2020, rma21, Miki2022, zhang2023learning}, offering advantages in search and rescue~\cite{tranzatto2022cerberus, miki2024learning}, planetary exploration~\cite{arm2023scientific, ding20222}, and hazardous area navigation~\cite{Wellhausen_anomaly}. The main challenge lies in ensuring their robustness and adaptability to diverse ground conditions like slippery surfaces, soft soils, and uneven terrains, necessitating control and navigation systems capable of sensing and responding to the environmental variations.

Physical simulators~\cite{leggedgym,raisim} are vital for training locomotion~\cite{Lee2020,Miki2022} and navigation~\cite{zhang2024resilient} policies for legged robots using reinforcement learning. They enhance controller robustness by simulating diverse terrains and conditions like rough or slippery surfaces~\cite{Miki2022,Wu2023-nz,Shi2023TerrainAwareQL,jenelten2024} and varying ground stiffness~\cite{Xu2023, choi2023learning}, enabling robots to adapt to different environments.
While randomizing physical parameters can improve the locomotion robustness, these parameters only become observable through direct interaction with the terrain~\cite{Lee2020, Miki2022}. 
It would be more beneficial for robots to anticipate conditions such as slippery surfaces or foot sinkage using exteroception before contact.
However, learning this in simulation is difficult. Although we can create photorealistic images, accurately replicating the physical effects of interacting with terrain in varied stiffness and friction is challenging. For example, simulating realistic grass visually is possible, but it's hard to match its real-world physical properties perfectly.
This mismatch between simulated and real-world terrain properties poses a challenge in transferring policies trained in simulation to the real world.
 
Self-supervised learning approaches allow training models to predict terrain-related properties from real-world images by labeling images from different sources~\cite{Frey2023,mattamala2024,Seo2023LearningOT,Castro2022HowDI, Wellhausen2019, Zurn2021}.
Recent approaches predict terrain-related property representations from vision, such as traversability~\cite{Frey2023,mattamala2024}, empirical ground reaction scores~\cite{Wellhausen2019}, or learned audio embeddings~\cite{Zurn2021}. These predictions are supervised using metrics like velocity tracking error, foot-terrain interaction force/torque, or vehicle-terrain interaction sounds.
Despite these models can predict various properties visually,
they cannot be directly used for training locomotion or navigation policies in simulation due to the challenge of accurately simulating signals like foot-terrain interaction force/torque or sound.

\begin{figure*}[t]
   \centering
   \includegraphics[scale=1.0]{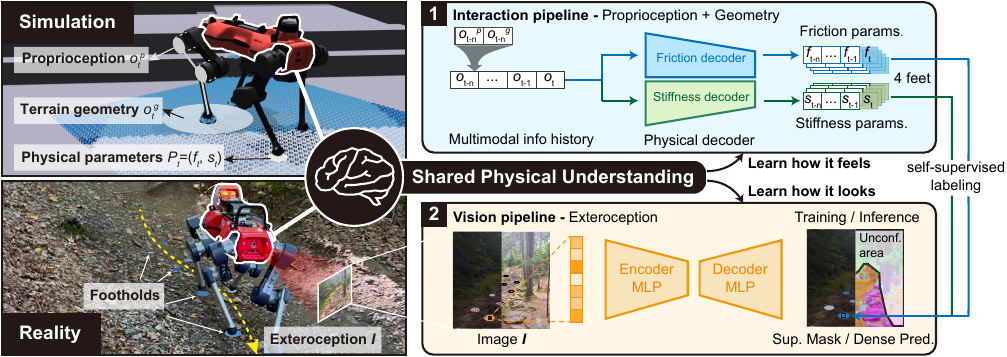}
   \caption{Overview of the two-stage self-supervised terrain physical parameter learning framework. A physical decoder in twin structure is trained in simulation to predict simulated friction and stiffness parameters per foot. The physical decoder transfers to the real world, where it provides self-supervised labels (within the supervision mask) to train a visual network on real-world image data. In the training stage, the visual network is trained with weak supervision only on the foothold pixels. In the inference phase, the visual pipeline processes all pixel features within an image and outputs the corresponding dense prediction of the simulated physical parameters with a confidence mask.}
   \label{main_figure}
   \vspace{-17pt}
\end{figure*}

In this work, we propose a framework (\figref{main_figure}) that aims to predict simulation parameters, instead of metrics like traversability, from vision using self-supervised learning.
% #These simulated parameters are defined in simulation and don’t exactly match real-world physical parameters. For example, fric. = 0.5 in simulation does not equal to a friction coefficient μ = 0.5 in the real world.
These simulated parameters are defined in simulation and don’t exactly match real-world physical parameters. For example, $\mathrm{stiffness} = 0.5$ in simulation does not equal a stiffness coefficient $k = 0.5$ in the real world.
By connecting real-world images with their corresponding physical properties in simulation, we can utilize these estimations for policy-training in simulation. The framework consists of two stages:
In stage one, a \emph{physical decoder} is trained in simulation to estimate physical parameters of the terrain from robotic proprioception and geometric sensing of the terrain. 
Using the simulated robotic interactions with different terrains, the physical decoder learns to estimate the simulated friction and stiffness that best approximate the behavior on various terrains.
In stage two, a \emph{visual network} is trained with real-world data to map visual features to simulated physical parameters using labels generated by the physical decoder.
The vision pipeline also uses anomaly detection to assess the reliability of the predicted terrain properties. A continuous learning process ensures that the robot can adapt to the complexity and uncertainty of real-world terrains, due to the fact that how it looks does not uniquely determine how it feels.

The main contributions of our work are as follows:
\begin{itemize}
    \item \textbf{Proposing a physical properties learning framework} aiming at transferring physical-parameter-aware locomotion and navigation policy to the real world.
    \item \textbf{The physical decoder}, a recurrent neural network with a gating mechanism trained in a contact-enhanced simulation, is able to predict terrain friction and stiffness parameters per foothold.
    \item \textbf{The visual network}, trained in a self-supervised manner to predict the simulated terrain parameters from the images, can be continuously updated and adapts quickly to new scenarios during real-world deployment.
    \item \textbf{Experimental analysis and quantitative evaluation} in simulation and real world demonstrating the superior performance of our physical decoder over the baseline method, as well as the efficacy of the visual pipeline.
\end{itemize}

% \hfill mds
%  
% \hfill August 26, 2015

\section{RELATED WORK}
\label{sec:related-work}
\subsection{Estimating Terrain Properties from Interaction}
Many studies have already shown that legged robots can efficiently estimate terrain-related properties through interaction.
Xu et al.~\cite{Xu2023} estimate friction and stiffness with sensor-measured normal and tangential contact forces acquired in formulaic pressing and rubbing motions for hexapod robots. 
Yu et al.~\cite{Yu2022ATW} explore the prediction of terrain roughness using hall effect sensors and a novel whisker-based system. 
Margolis et al.~\cite{active} predict friction and roughness using a neural network with an active-sensing policy incorporating probing motions.
These prior works either require additional sensors or rely on probing motions which hinder the robot from moving freely. 
Miki et al. \cite{Miki2022} train a recurrent belief state decoder in simulation, which is a part of the locomotion policy, to predict foot contact information, including friction coefficient. The belief decoder is implemented by a recurrent architecture, which processes historical sensor data, and shows its capability to estimate friction when the robot steps onto a slippery platform without relying on any specific motions or sensors.
Inspired by this design, we train a network to predict friction and stiffness accurately for each foot.

\subsection{Estimating Terrain Properties from Vision}
Identifying terrain properties through interaction is well studied, but poses potential risks, such as being slippery or stuck due to excessive foot sinkage. In contrast, vision-based methods offer the opportunity to estimate these properties from semantic-rich images and help avoid these hazards in advance.
Multiple works explore extracting meaningful feature embeddings from image data \cite{dinov1,dinov2}.
Correlating semantic information to those feature embeddings relying on pre-trained models has been shown to be specifically effective with limited available training data \cite{Frey2023,Zurn2021,active}.

A common approach to obtain labels for image training is to use self-supervision in hindsight and cross-modality~\cite{Wellhausen2019, Zurn2021, Frey2023, mattamala2024}. Wellhausen et al.~\cite{Wellhausen2019} train a \gls{cnn} to associate a learned terrain property metric named \textit{ground reaction score} with camera images. Margolis et al.~\cite{active} used the predicted values from a physical estimator as labels to train a linear layer on the dense feature representation. Lee et al. \cite{Anqiao23} reproject footholds to learn the geometry of the support surface from image data.
Frey and Mattamala et al.~\cite{Frey2023, mattamala2024} reproject the footprint of the robot labeled with the velocity tracking error to approximate traversability and use it as a label in the image space. In this work, we use a similar self-supervision principle with a label-projection method as~\cite{Wellhausen2019, Anqiao23}, adapting per-foot labeling.

Even though pre-trained features offer good generalization, models trained on limited data may not adapt to entirely new environments. Following~\cite{Frey2023,mattamala2024}, we adapt the network to the deployment environment online during the mission and predict the physical parameters of the environment, allowing to transfer physical-parameter-aware locomotion and navigation policies to the real world.

\subsection{Anomaly Detection}
Anomaly detection methods enable identification of the input data that is \gls{ood}. Most approaches model the input data distribution and use a similarity measure to determine if a sample is \gls{ood}, based on a predefined threshold. 
Richter and Roy~\cite{richter} train an autoencoder to reconstruct image data and use the reconstruction loss threshold to identify \gls{ood} samples. 
Wellhausen et al.~\cite{Wellhausen_anomaly} use a loss threshold with a normalizing flow network. Frey et al.~\cite{Frey2023} follow~\cite{richter} and dynamically adjust the anomaly detection threshold by fitting a Gaussian distribution to positive samples. We extend~\cite{Frey2023}, however, fit a Gaussian Mixture Model (GMM) with two components that yield a hyperparameter-free threshold.

\section{METHOD}
\label{sec:method}
The framework (\figref{main_figure}) for online and self-supervised physical environment understanding consists of two stages. The physical decoder provides terrain parameter labels for the online training of the visual network; The visual pipeline outputs confidence-masked dense predictions of the physical properties of the environment in image space. 

The \emph{physical decoder} (Sec. \ref{subsec:physical-decoders}) has a twin network to estimate the friction and stiffness values of each foothold, respectively. These footholds, along with their associated friction and stiffness values, are then projected into the camera image, where they serve as labels for training the visual network. The visual network uses a \gls{mlp} to predict the physical parameters (Sec. \ref{subsec:visual-decoder}) from pixel-wise feature embeddings, and a confidence mask is generated using anomaly detection (Sec. \ref{subsec:ood-detection}). A \textit{Mission Graph} and a \textit{learning thread} (Sec. \ref{subsec:online-training}) are established for online dataset storage and training of the visual network. 

\subsection{Physical Decoder} 
\label{subsec:physical-decoders}
\begin{figure}[t]
   \centering
   \includegraphics[scale=1.0]{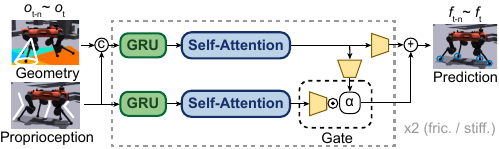}
   \caption{Physical decoder architecture in the form of a twin network. Friction and stiffness are predicted by each separate network. The yellow trapezoidal blocks are \glspl{mlp}.}
   \label{phy}
   \vspace{-10pt}
\end{figure}

As illustrated in \figref{main_figure}, we train a physical decoder capable of predicting both terrain friction and stiffness on a per-foot basis using proprioceptive and exteroceptive information within a simulation environment. The decoder learns the correspondence between the simulated terrain parameters and the resulting motion in the simulation, which is then intended to be transferred to real-world applications.
The decoder consists of a twin network architecture, as demonstrated in \figref{phy}. Both networks share the same input observation history $O = \{o_{t-n}, ..., o_{t}\}$, where $n$ is the history length. Each observation $o_{t}$ comprises the robotic proprioception $o_t^p$, including command, joint, body information as well as leg phase information, along with the terrain geometry $o_t^{g}$ represented by height samples around each foot in a circular sampling pattern, as~\cite{Miki2022}. 
Each decoder predicts a 4-dimensional vector representing the friction or stiffness per foothold. When the foot is in the swing phase, the corresponding friction or stiffness is determined by the terrain parameter within the foot's projection area on the ground.  
Inspired by the recurrent structures in~\cite{Miki2022}, each twin network comprises \glspl{gru} blocks, self-attention layers, and \gls{mlp} prediction heads with an additional gating mechanism. 
The \glspl{gru} retain relevant historical information for friction or stiffness estimation, which is further refined by the self-attention layers.
Considering two different modalities of input information, the network architecture consists of a proprioception-only and concatenated-information path. 
In each path, a \gls{mlp} predicts the physical terrain properties. 
The gating mechanism allows to weight the contribution of each path.
Unlike~\cite{Miki2022}, where terrain property decoding is treated as an auxiliary task, we optimize the physical decoder independently from the locomotion policy.
Additional details on the network architecture, training environment, data collection, and training hyperparameters are provided in \secref{subsec:implementation-details}.

\subsection{Visual Network}
\label{subsec:visual-decoder}
\begin{figure}[t]
   \centering
   \includegraphics[scale=1.0]{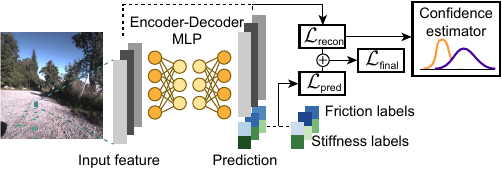}
   \caption{Visual network architecture and losses used for training. The decoder is in an encoder-decoder structure for the simultaneous \gls{ood} detection and physical parameters regression. Friction and stiffness values of each pixel feature in the input image are predicted at the same time.}
   \label{visdecoder}
   \vspace{-5pt}
\end{figure}

We implement the visual network in an encoder-decoder structure (\figref{visdecoder}). We first extract pixel-wise feature embeddings using a pre-trained DINOv2 backbone~\cite{dinov2}, compared to DINOv1~\cite{dinov1} used in~\cite{Frey2023}. 
Since friction and stiffness aren't strongly dependent on terrain geometry, using powerful image features extracted by DINOv2 containing implicit geometric and other information is enough to build a good mapping between these parameters and features without explicit depth sensing.
In addition, we follow~\cite{mattamala2024}, an extension of~\cite{Frey2023}, and predict the terrain properties per-pixel instead of using SLIC superpixels~\cite{slic} to aggregate feature embeddings. 

The visual network outputs the reconstructed input feature for anomaly detection (Sec. \ref{subsec:ood-detection}) in confidence estimator and friction and stiffness predictions per pixel. The total loss per pixel during training is a weighted combination of reconstruction loss $\mathcal{L}_\mathrm{recon}$ and regression loss $\mathcal{L}_\mathrm{pred}$ given by:
\begin{equation}
\mathcal{L}_\mathrm{recon} = \text{MSE}(x_\mathrm{recon}, x_\mathrm{label})
\end{equation}
\begin{equation}
\mathcal{L}_\mathrm{pred} = \text{MSE}(y_\mathrm{pred}, y_\mathrm{label})
\end{equation}
\begin{equation}
\mathcal{L}_\mathrm{final} = w_\mathrm{recon} \cdot \mathcal{L}_\mathrm{recon} +
w_\mathrm{pred} \cdot \mathcal{L}_\mathrm{pred},
\end{equation}
where \( w_\mathrm{recon} \) and \( w_\mathrm{pred} \) are the
weighting coefficients for the reconstruction and prediction loss respectively.
More implementation details on the label generation for the visual network training, and the underlying graph storage implementation are provided in Sec. \ref{subsec:online-training}. 
\vspace{-0.3cm}
\subsection{Anomaly Detection}
\label{subsec:ood-detection}
\begin{figure}[t]
   \centering
   \includegraphics[scale=1.0]{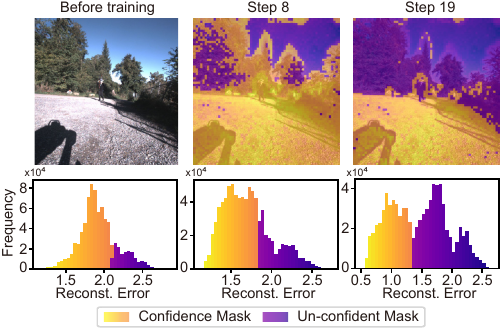}
   \caption{Evolution of the reconstruction loss distribution with the increase of training steps. The reconstruction loss distribution is unimodal before training, while changed to a bimodal distribution during training. Yellow-orange indicates In-Distribution (ID) data, while purple-blue is for Out-Of-Distribution (OOD) data.}
   \label{hists}
   \vspace{-5pt}
\end{figure}

When deploying legged robots, they interact only with a small part of the scene. Our self-supervised labeling method, therefore, can only provide physical parameter supervision for regions with footholds. However, the visual network predicts parameters for the full image including irrelevant areas (e.g. sky and trees). This directly necessitates reasoning about the reliability of these predictions. 
We implement an anomaly detection strategy using the encoder-decoder architecture of the visual network to predict \gls{ood} regions, by learning to reconstruct the feature embedding of pixels with associated physical properties. 
Following the implementation of~\cite{Frey2023}, we observed a bimodal reconstruction error distribution during training as shown in \figref{hists}.

To dynamically adjust the threshold for \gls{ood} discrimination, we use a \gls{gmm}~\cite{GMM} with $k=2$ and \emph{full} covariance type to fit the bimodal distribution. 
Fitting a \gls{gmm} to the histogram is independent of the training or inference process of the visual network and the threshold can be dynamically adapted per image. 
This process requires no hyper-parameter tuning across datasets and we will show its advantage in Sec. \ref{subsec:visual-decoder-exp}.

\begin{figure}[t]
  \centering
  \includegraphics[scale=1.0]{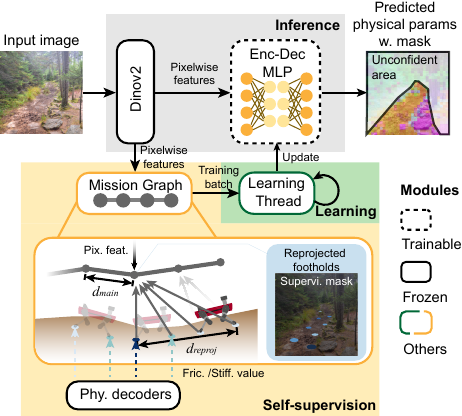}
  \caption{Online training framework adapted from~\cite{Frey2023}. The inference task extracts features and outputs masked dense predictions. The self-supervision task contains the Mission Graph to store paired input features and labels provided by the physical decoder. The learning task performs continuous training of the visual network in the learning thread. 
  }
  \label{vision_pipeline}
  \vspace{-5pt}
\end{figure}
\vspace{-0.3cm}
\subsection{Online Training}
\label{subsec:online-training}
The visual network is trained online during the deployment to overcome the generalization limitations of offline training on limited datasets. It can update the mapping between visual features and physical parameters after it encounters a different pattern during environmental interaction.
We perform three tasks in parallel during the mission, as shown in \figref{vision_pipeline}. \emph{Task~1}: fast prediction of terrain properties based on the received image; \emph{Task~2}: self-supervised labeling of training data by associating footholds (only when in contact) with physical terrain parameters to camera images; \emph{Task~3}: training the visual network given the labeled data. 
All three tasks are connected by the Mission Graph, the central hub, allowing to receive, store, and transfer training-relevant data. 

\subsubsection{Mission Graph}
\label{subsubsec:mission-graph}
The Mission Graph accumulates training data by storing image features and the associated physical parameters. To facilitate associating image features and terrain parameters, each mission node stores the dense feature map, a supervision mask of footholds with associated friction and stiffness, of the same height and width as the camera image, and the associated camera intrinsics and extrinsics. 

\subsubsection{Inference Task}
\label{subsubsec:inference-task}
This task processes camera images by extracting dense features using DINOv2. 
The visual network predicts friction and stiffness values using the dense pixel-wise features and outputs a confidence mask for each image.

\subsubsection{Self-supervision Task}
\label{subsubsec:ss-thread}
The self-supervision task receives the generated dense features from the inference task with the associated camera intrinsic and extrinsic, as well as the labeled footholds with physical parameters from the physical decoder.
Dense features are stored in a new node within the Mission Graph if and only if the camera image is recorded at a position farther away than \( d_{\text{main}} \), with respect to the latest added node. When a new foothold with labeled physical parameters is received, the foothold is projected on all camera images stored within a range of \( d_{\text{reproj}} \) in the Mission Graph using the same projection method in~\cite{Wellhausen2019}. The projection is used to fill the supervision masks with the friction and stiffness values. This establishes the pairing of feature embeddings to terrain physical parameters. 

\subsubsection{Learning Task}
\label{subsubsec:learning-thread}
In each iteration of the learning process, a batch of mission nodes containing a supervision signal are randomly selected. The resulting training batch is used to update the visual network using gradient-based optimization.
\vspace{-0.3cm}
\subsection{Implementation Details}
\label{subsec:implementation-details}

\subsubsection{Interaction pipeline}
\label{subsubsec:interaction-pipeline}
In the physical decoder architecture, the 1-layer \glspl{gru} have a 100-dimensional hidden state. 
The number of hidden and output units of the prediction head \glspl{mlp} is [64,32,4], while it is [64,64,4] for the \gls{mlp} controlling the sigmoid gating unit. All \glspl{mlp} use LeakyReLU activation functions.
The history length of input observation sequences is 50, and input sequences are zero-padded for missing historical information during both training and testing. During training, the hidden states of the \glspl{gru} are reset to zeros for each new sequence.

For data collection, all the data for training and testing the physical decoder is collected in legged gym~\cite{leggedgym} enhanced with procedurally generated terrain~\cite{miki2024learning} and an additive soft terrain contact model~\cite{choi2023learning}. The training and testing environments are randomly generated per terrain patch with varying geometry, stiffness, and friction parameters. The friction value ranges from 0 to 1, and stiffness from 1 to 10. The legged robots are controlled by a robust locomotion policy~\cite{Miki2022} in simulation, and their command velocities are uniformly sampled between $\SI{-1.3}{m/s}$ and $\SI{1.3}{m/s}$ with an angular velocity between $\SI{-0.3}{rad/s}$ and $\SI{0.3}{rad/s}$. We collect a dataset consisting of $\SI{18.3}{h}$ and $\SI{5}{h}$ real-time equivalent locomotion in simulation for training and testing.  

For training hyper-parameters, we use a batch size of 64. Label normalization is dynamically performed by continuously updating the running mean and standard deviation when new data is ingested. Furthermore, labels are weighted by their interval-based frequencies when calculating the loss, ensuring a balanced training process. For the optimizer, we use Adam \cite{kingma2014adam} with a learning rate of 0.001 and a weight decay factor of 0.00001. We train for a total of 100 epochs, which takes around 30 minutes on a single GPU. 

\subsubsection{Vision pipeline}
\label{subsubsec:vision-pipeline}
In the visual network's encoder-decoder architecture, hidden layers are configured as [128,32,128,384+2], where the output of 384 corresponds to the input feature dimension, while 2 additional channels are used for the friction and stiffness prediction. In preprocessing, we first scale the camera image to a height of 1078 and then center-crop it to 910$\times$910 pixels. The loss weights are set to: \( w_{pred}=0.1 \) and \( w_{recon}=0.9 \). For the Mission Graph, the default outdoor setting is \( d_{\text{main}}=\SI{1}{m} \), \( d_{\text{reproj}}=\SI{5}{m} \), and $ d_{\text{main}}$ is set to \SI{0.2}{m} for indoor experiments. For online training, we use the Adam optimizer with a learning rate of 0.001, weight decay factor of 0.001, and batch size of 100. The training loop runs at around \SI{2.0}{Hz}.
For key hyperparameters, we conducted a grid search and chose the best based on the validation dataset.
\section{EXPERIMENTS}
\label{sec:experiments}
We test the interaction-based and vision-based components of our terrain physical parameter learning framework in effectiveness and accuracy consecutively. 
First, we evaluate the prediction performance of the physical decoder in simulation (\secref{subsubsec:test-in-simulation}). 
Second, we compare prediction results for friction and stiffness against a baseline when walking over a slippery whiteboard and a foam board in the real world (\secref{subsubsec:test-in-realworld}). 
Third, we show a quantitative analysis of our model with a digital-twin-based experiment, where the predicted friction value in the real world results in a similar motion when rolling out the same action in simulation (\secref{subsec:digital-twin}). It justifies the usage of our physical decoder to generate labels for the self-supervised training of the visual network. 
Next, we evaluate our online-trained vision pipeline in an indoor environment (\secref{subsec:visual-decoder-exp}). 
Lastly, we compare different methods for anomaly detection. 
All experiments are conducted using an ANYmal D quadruple robot with a robust locomotion policy~\cite{Miki2022} and state estimation given by~\cite{bloesch2013state}.  
Experiments were run on an Nvidia RTX 4080 GPU with Intel i7-12700H CPU. 
\vspace{-0.3cm}
\subsection{Physical Decoder}
\label{subsec:physical-decoders-exp}
\subsubsection{Test in simulation}
\label{subsubsec:test-in-simulation}
We test the effectiveness of our trained physical decoder in simulation and quantitatively evaluate its performance compared to a \emph{Baseline} method \cite{Miki2022} for friction estimation.
\begin{figure}[t]
   \centering
   \includegraphics[scale=1.0]{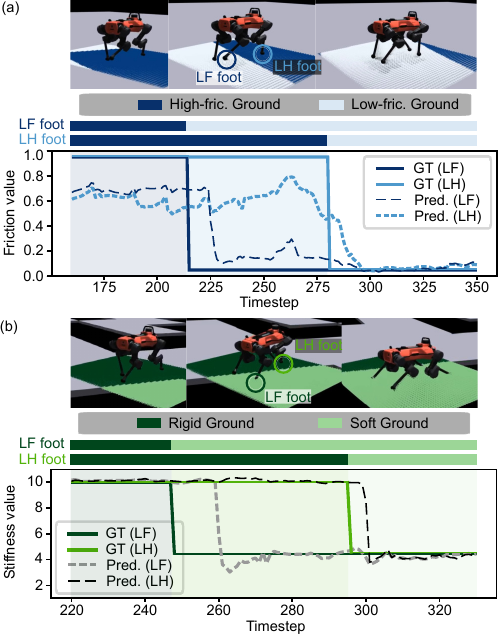}
   \caption{Friction and stiffness estimation in simulation. One steps equals \SI{20}{ms}. Prediction results shown here are from the left-front (LF) and left-rear (LH) foot.}
   \label{phy_sim}
\end{figure}
As shown in \figref{phy_sim}a, the friction prediction of the front and hind feet transits quickly and correctly over time when the robot walks from a high-friction ($0.95$) region to a low-friction ($0.05$) one. 
The friction prediction of the front foot drops firstly at step 225, while the decrease in friction for the hind foot happens around 60 steps later. 
This indicates that our friction network is able to identify parameters exactly per foot contact, rather than an average measure of four foot contacts.
Within the high friction region, we can observe that the friction prediction around $0.6$ doesn't perfectly match with the ground truth value of $0.95$. This is due to the fact that the motion on high friction terrains ($0.6~1.0$) is very similar given that no foot slippage occurs, which renders the exact friction value unobservable. 
Even though the prediction in high friction range is less precise, it is acceptable and sufficient to adapt the locomotion or navigation policy to varying frictions.

\begin{figure}[t]
   \centering
   \includegraphics[scale=1.0]{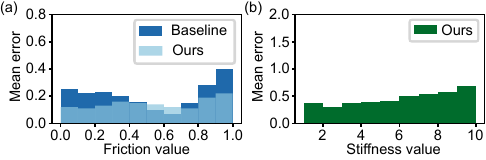}
   \caption{Simulation test error histograms of friction (a) and stiffness (b) prediction. The Baseline is the decoder in~\cite{Miki2022}.}
   \label{histboth}
\end{figure}

For the stiffness prediction, the decoder is able to estimate the terrain stiffness accurately across the full range from 1 to 10, as shown in \figref{phy_sim}b. In both the friction and stiffness testing scenarios, we can observe a delay of around 20 timesteps (\SI{0.4}{s}) between the ground truth and decoder prediction. This is due to the fact that the foot enters the low friction/stiffness region but has not interacted with the new terrain patch. 
The statistics of the prediction error in \gls{mae} histogram over the 5-hour test data in terms of friction and stiffness are illustrated in \figref{histboth}.
Regarding friction prediction, it shows that our method outperforms \emph{Baseline} on average in terms of \gls{mae} (0.21	$\rightarrow$ 0.15).
For high friction ranges, both methods cannot accurately distinguish different friction values. 
For stiffness prediction, our method can predict stiffness with an overall \gls{mae} of 0.46. 
This indicates that stiffness can be predicted precisely ($\sim$\SI{5}{\%} deviation), given the large stiffness range of 1 to 10.
Notably, we observed experimentally that the gating mechanism tends to emphasize the path containing solely proprioceptive information for friction prediction, while relying more on the path containing exteroception for stiffness prediction. 
This observation can be attributed to the fact that stiffness is observable when the foot penetrates the ground, which can be detected based on the exteroception, whereas friction predictions are less dependent on the terrain geometry.

\subsubsection{Test in real world}
\label{subsubsec:test-in-realworld}
\begin{figure}[t]
   \centering
   \includegraphics[scale=1.0]{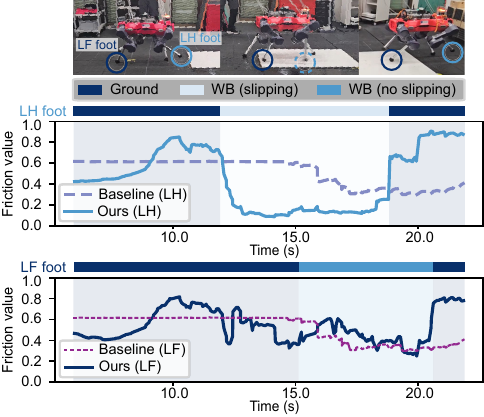}
   \caption{Physical decoder friction estimation. Only left-front (LF) and left-rear (LH) feet are shown in the plot. The robot walks backwards.
   }
   \label{realfric}
\end{figure}
\vspace{-0.3cm}
\begin{table}[h]
\centering
\caption{Mean Friction Prediction Error (LH foot)}
\begin{tabular}{ccc}
\hline
\textbf{Scene}  & \textbf{Baseline} & \textbf{Ours} \\
\hline
WB & $0.33 \pm 0.11$ & $\textbf{0.02} \pm 0.08$ \\
GROUND & $\textbf{0.00} \pm 0.00$ & $\textbf{0.00} \pm 0.00$ \\
\hline
\end{tabular}
\label{tab:pred_error_real}
\end{table}
\vspace{-0.3cm}
\begin{figure}[t]
   \centering
   \includegraphics[scale=1.0]{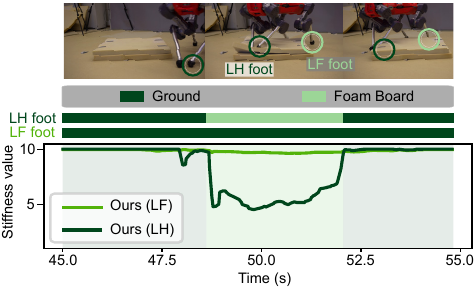}
   \caption{Physical decoder stiffness estimation. Left-front (LF) and left-rear (LH) feet are shown in the plot. 
   }
   \label{realstiff}
\end{figure}
To demonstrate that our in-simulation trained physical decoder transfers effectively to the real world, we conducted separate experiments for varying friction and stiffness. 
For friction, we tele-operated the robot to move backwards from a high-friction ground (\emph{GROUND}) onto a slippery, water-covered whiteboard (\emph{WB}).
As shown in \figref{realfric}, our friction decoder accurately distinguishes between different friction levels per foot (e.g., rear feet on the slippery area, front feet on high-friction ground), while the \emph{Baseline}~\cite{Miki2022} predicts the same estimate for all feet.
For stiffness, we had the robot walk over a soft foam board, as illustrated in \figref{realstiff}. 
Our stiffness decoder identifies the low-stiffness area when stepping on the foam and quickly adjusts to a high value upon returning to rigid ground.
\vspace{-0.45cm}
\subsection{Digital Twin}
\label{subsec:digital-twin}
\begin{figure}[t]
   \centering
   \includegraphics[scale=1.0]{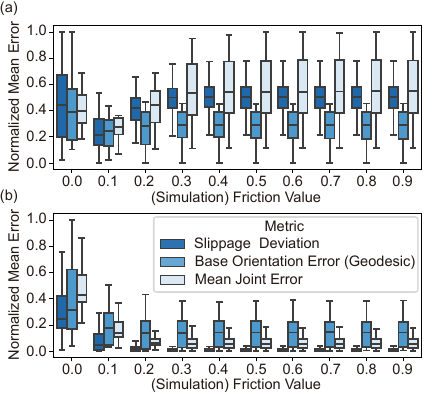}
   \caption{Boxplot result of the digital twin experiment. (a) Error statistics of the WB experiment. (b) Error statistics of the GROUND experiment. }
   \label{digiboxplot}
   \vspace{-0.3cm}
\end{figure}
Although the efficacy and qualitative sim-to-real transferability of the physical decoder have been demonstrated in previous experiments, the numerical correctness of the predicted simulation parameters hasn't been verified in the real world. By ensuring these parameters yield comparable motions in simulations, we can render the physical properties of the terrain observable for locomotion or navigation policies trained in simulation with privileged information.

To calculate the numerical correctness of the predicted simulation parameters in the real world, we design a digital twin experiment, inspired by \cite{Haddeler2022RealtimeDD}. It can align simulation parameters with the real world by comparing real and simulated robot motions under various physical conditions, leveraging data from the friction experiment (\secref{subsubsec:test-in-realworld}). To quantify motion similarity, we introduce three metrics: simulated joint position error (absolute error), base orientation error (geodesic distance), and foot slippage distance error, which is calculated from the movement of the foot from ground contact to lift-off.

We initialize the simulated robot with the same base orientation, joint positions, and velocities as recorded from the real robot. 
Orientation roll and pitch are reliably captured by the onboard IMU, while the base height is determined by assuming one foot is always on the ground. 
The simulation terrain is reconstructed to match the terrain geometry in the corresponding real-world experiment. An accurate reconstruction is the key to provide discernible comparison results. Due to the current limitation of the simulation and the manual reconstruction pipeline, this method currently supports only flat, rigid surfaces.
The recorded dataset is split into snipppets: whiteboard (\emph{WB}) and high-friction ground (\emph{GROUND}). We selected 18 \emph{WB} snippets, each 20 timestamps (\SI{400}{ms}) long, where foot slippage occurred, and 220 \emph{GROUND} snippets of the same length with no slippage. 
The lower number of \emph{WB} samples is due to the rarity of slippage events. 
We then simulate \SI{400}{ms} rollouts for 10 friction parameters ranging from 0.0 to 0.9, noting that longer trajectories tend to diverge from reality. 
All metrics are averaged per snippet. 

The results shown in \figref{digiboxplot} indicate that the lowest error range for \emph{WB} is within a range of $(0.0,\;0.2)$ and a range of $(0.3,\;1.0)$ for \emph{GROUND}. 
This result allows us to reinterpret \figref{realfric}, where we can observe high friction predictions $(>0.4)$ on \emph{GROUND} and low friction estimates $(\approx 0.1)$ on \emph{WB}. We use these two ranges above for quantitative analysis of the LH foot in \figref{realfric}. The prediction error per timestep is set to zero if the predicted value is within the range. Otherwise, the error is given by the absolute error to the nearest boundary of the prediction range. The quantitative result in \tabref{tab:pred_error_real} shows that our approach outperforms the \emph{Baseline}, particularly with \emph{WB}, and performs comparably on \emph{GROUND} due to its broad valid range.
The results indicate that the physical decoder trained in simulation can successfully transfer to the real world under controlled conditions. Given the design of our simulation environment to closely mimic real-world physical interactions, we assume the shown results also generalize to stiffness. However further validation is still needed after we enhance the digital twin method in the future.
\vspace{-0.3cm}
\subsection{Visual Network}
\label{subsec:visual-decoder-exp}
We train the visual network in a self-supervised manner on similar robot data as the one in the digital twin experiment (\secref{subsec:digital-twin}). The training data consists of 22 images.
Within the experiment, we aim to evaluate the terrain property prediction and the confidence prediction based on our adaptive anomaly threshold. 

\begin{figure}[t]
   \centering
   \includegraphics[scale=1.0]{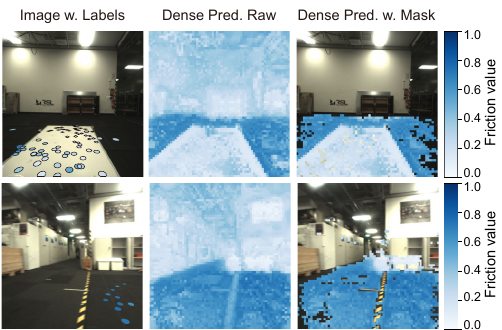}
   \vspace{-0.2cm}
   \caption{Dense friction prediction frames. We also output the mean value of friction prediction in the whiteboard/ground areas (Row 1: mask acc. 0.84, Whiteboard pred mean 0.22, Ground pred mean 0.60; Row 2: mask acc. 0.80, Ground pred mean 0.62).}
   \label{vid}
   \vspace{-0.3cm}
\end{figure}

An example prediction of our vision pipeline is shown in \figref{vid}. To evaluate the confidence mask prediction, we manually label the ground and whiteboard areas within the images as the ground truth mask. The \gls{iou} metric for the full dataset is reported in \tabref{tab:dense_pred_err}. Our method accurately distinguishes between the floor (including the \emph{WB} and \emph{GROUND} area) and the remaining part of the scene, where most of the predicted physical properties are unreliable, achieving an \gls{iou} of 0.82.

From the previous digital twin experiment, we establish a friction range for the \emph{WB} of $(0.0,0.2)$ and $(0.3,1.0)$ for the \emph{GROUND}. 
The same rule as in \secref{subsec:digital-twin} applies to the error calculation for each pixel prediction. The mean error for the \emph{WB} area is 0.03 across all images and 0.0 for the \emph{GROUND}, given the large interval of valid friction values. 

\begin{figure}[t]
   \centering
   \includegraphics[scale=1.0]{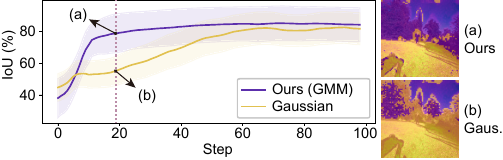}
   \caption{Confidence mask accuracy development as the increase of training step. Two insets on the right show reconstruction errors visualized in the same image as in \figref{hists} on the 19th step.}
   \label{abl1}
   \vspace{-5pt}
\end{figure}

In addition, we ablated the choice of our proposed \gls{gmm} method, compared to the original method~\cite{Frey2023}. 
We deployed our physical terrain property pipeline on an outdoor dataset, consisting of 82 images, and report the anomaly detection performance. 
The result in Fig. \ref{abl1}, with two example images, shows that our method can achieve a higher mask accuracy more rapidly than our previous method. 

More experimental results of our framework in off-road scenarios and environments with changing physical properties can be found in the supplementary video  \url{https://bit.ly/3Xo5AA8}.
\vspace{-0.3cm}
\begin{table}[h]
\centering
\caption{Dense Prediction Error across the full dataset.}
\begin{tabular}{ccc}
\hline
\textbf{Scene} & \textbf{Fric. pred. error} & \textbf{Conf. mask acc. (\gls{iou})}  \\
\hline
WB & $0.03 \pm 0.05$ & \multirow{2}{*}{$0.82 \pm 0.05$} \\
GROUND & $0.00 \pm 0.01$&  \\
\hline
\end{tabular}

\label{tab:dense_pred_err}
\end{table}
\vspace{-0.3cm}

\section{CONCLUSIONS AND DISCUSSIONS}
We successfully trained our physical decoder to estimate terrain friction and stiffness from the interaction, outperforming the baseline method with MAE of 0.15 and 0.46 separately. 
In addition, we showed that the identified parameters of the physical decoder in the real world align with the simulation parameters in the digital twin experiment. Lastly, we demonstrated our online vision pipeline effectively predicts masked terrain friction from vision without interaction.

However, further improvements in terms of prediction stability and inaccuracies are needed. For example, when the robot stands, we observed non-stationary friction estimates, or the change in stiffness for a thin \SI{5}{cm} foam board in the real world could not be correctly predicted. This is likely due to insufficient training data or noisy real-world geometric observations. We expect by carefully incorporating noise during training our model can be further improved and robustified. Besides, we can enhance the digital-twin-based method to support extensive quantitative analysis of more complex real-world scenarios, beyond a flat and rigid surface.

Our work opens up the possibility of transferring physical-terrain-parameter-aware locomotion and navigation policies, trained in simulation,
to the real world, utilizing both visual prediction and proprioception.
In the future, we aim to address the limitations and train such policies to make legged robots more capable in the real world.

% needed in second column of first page if using \IEEEpubid
%\IEEEpubidadjcol

% Note that the IEEE does not put floats in the very first column
% - or typically anywhere on the first page for that matter. Also,
% in-text middle ("here") positioning is typically not used, but it
% is allowed and encouraged for Computer Society conferences (but
% not Computer Society journals). Most IEEE journals/conferences use
% top floats exclusively. 
% Note that, LaTeX2e, unlike IEEE journals/conferences, places
% footnotes above bottom floats. This can be corrected via the
% \fnbelowfloat command of the stfloats package.

% Can use something like this to put references on a page
% by themselves when using endfloat and the captionsoff option.
\ifCLASSOPTIONcaptionsoff
  \newpage
\fi

% trigger a \newpage just before the given reference
% number - used to balance the columns on the last page
% adjust value as needed - may need to be readjusted if
% the document is modified later
%\IEEEtriggeratref{8}
% The "triggered" command can be changed if desired:
%\IEEEtriggercmd{\enlargethispage{-5in}}

% references section

% can use a bibliography generated by BibTeX as a .bbl file
% BibTeX documentation can be easily obtained at:
% http://mirror.ctan.org/biblio/bibtex/contrib/doc/
% The IEEEtran BibTeX style support page is at:
% http://www.michaelshell.org/tex/ieeetran/bibtex/
%\bibliographystyle{IEEEtran}
% argument is your BibTeX string definitions and bibliography database(s)
%\bibliography{IEEEabrv,../bib/paper}
%
% <OR> manually copy in the resultant .bbl file
% set second argument of \begin to the number of references
% (used to reserve space for the reference number labels box)
\vspace{-0.3cm}
\footnotesize
\bibliography{references}

\begin{thebibliography}{10}

\bibitem{Lee2020}
Joonho Lee and et~al.
\newblock Learning quadrupedal locomotion over challenging terrain.
\newblock {\em Science robotics}, 5(47):eabc5986, 2020.

\bibitem{rma21}
Ashish Kumar, Zipeng Fu, Deepak Pathak, and Jitendra Malik.
\newblock {RMA:} rapid motor adaptation for legged robots.
\newblock In Dylan~A. Shell, Marc Toussaint, and M.~Ani Hsieh, editors, {\em Robotics: Science and Systems XVII, Virtual Event, July 12-16, 2021}, 2021.

\bibitem{Miki2022}
Takahiro Miki and et~al.
\newblock Learning robust perceptive locomotion for quadrupedal robots in the wild.
\newblock {\em Science Robotics}, 7(62):eabk2822, 2022.

\bibitem{zhang2023learning}
Chong Zhang and et~al.
\newblock Learning agile locomotion on risky terrains.
\newblock {\em arXiv preprint arXiv:2311.10484}, 2023.

\bibitem{tranzatto2022cerberus}
Marco Tranzatto and et~al.
\newblock Cerberus in the darpa subterranean challenge.
\newblock {\em Science Robotics}, 7(66):eabp9742, 2022.

\bibitem{miki2024learning}
Takahiro Miki and et~al.
\newblock Learning to walk in confined spaces using 3d representation.
\newblock {\em arXiv preprint arXiv:2403.00187}, 2024.

\bibitem{arm2023scientific}
Philip Arm and et~al.
\newblock Scientific exploration of challenging planetary analog environments with a team of legged robots.
\newblock {\em Science robotics}, 8(80):eade9548, 2023.

\bibitem{ding20222}
Liang Ding, Ruyi Zhou, Ye~Yuan, and et~al.
\newblock A 2-year locomotive exploration and scientific investigation of the lunar farside by the yutu-2 rover.
\newblock {\em Science Robotics}, 7(62):eabj6660, 2022.

\bibitem{Wellhausen_anomaly}
Lorenz Wellhausen and et~al.
\newblock Safe robot navigation via multi-modal anomaly detection.
\newblock {\em IEEE Robotics and Automation Letters}, 5(2):1326–1333, April 2020.

\bibitem{leggedgym}
Nikita Rudin and et~al.
\newblock Learning to walk in minutes using massively parallel deep reinforcement learning.
\newblock In {\em Proceedings of the 5th Conference on Robot Learning}, volume 164, pages 91 -- 100, 2022.

\bibitem{raisim}
Jemin Hwangbo and et~al.
\newblock Per-contact iteration method for solving contact dynamics.
\newblock {\em IEEE Robotics and Automation Letters}, 3(2):895--902, 2018.

\bibitem{zhang2024resilient}
Chong Zhang and et~al.
\newblock Resilient legged local navigation: Learning to traverse with compromised perception end-to-end.
\newblock In {\em 41st IEEE Conference on Robotics and Automation (ICRA 2024)}, 2024.

\bibitem{Wu2023-nz}
Jinze Wu, Guiyang Xin, Chenkun Qi, and Yufei Xue.
\newblock Learning robust and agile legged locomotion using adversarial motion priors.
\newblock {\em IEEE Robotics and Automation Letters}, 8(8):4975--4982, August 2023.

\bibitem{Shi2023TerrainAwareQL}
Hao bin Shi and et~al.
\newblock Terrain-aware quadrupedal locomotion via reinforcement learning.
\newblock {\em arXiv preprint arXiv:2310.04675}, 2023.

\bibitem{jenelten2024}
Fabian Jenelten, Junzhe He, Farbod Farshidian, and Marco Hutter.
\newblock Dtc: Deep tracking control.
\newblock {\em Science Robotics}, 9(86):eadh5401, 2024.

\bibitem{Xu2023}
Peng Xu and et~al.
\newblock Learning physical characteristics like animals for legged robots.
\newblock {\em National Science Review}, 10, 4 2023.

\bibitem{choi2023learning}
Suyoung Choi and et~al.
\newblock Learning quadrupedal locomotion on deformable terrain.
\newblock {\em Science Robotics}, 8(74):eade2256, 2023.

\bibitem{Frey2023}
Jonas Frey and et~al.
\newblock {Fast Traversability Estimation for Wild Visual Navigation}.
\newblock In {\em Proceedings of Robotics: Science and Systems}, July 2023.

\bibitem{mattamala2024}
Matias Mattamala and et~al.
\newblock Wild visual navigation: Fast traversability learning via pre-trained models and online self-supervision.
\newblock {\em under review for Autonomous Robots}, 2024.

\bibitem{Seo2023LearningOT}
Junwon Seo and et~al.
\newblock Learning off-road terrain traversability with self-supervisions only.
\newblock {\em IEEE Robotics and Automation Letters}, 8:4617--4624, 2023.

\bibitem{Castro2022HowDI}
Mateo~Guaman Castro and et~al.
\newblock How does it feel? self-supervised costmap learning for off-road vehicle traversability.
\newblock {\em 2023 IEEE International Conference on Robotics and Automation (ICRA)}, pages 931--938, 2022.

\bibitem{Wellhausen2019}
Lorenz Wellhausen and et~al.
\newblock Where should {I} walk ({P}redicting terrain properties from images via self-supervised learning).
\newblock {\em IEEE Robotics and Automation Letters}, 4:1509--1516, 4 2019.

\bibitem{Zurn2021}
Jannik Z{\"u}rn, Wolfram Burgard, and Abhinav Valada.
\newblock Self-supervised visual terrain classification from unsupervised acoustic feature learning.
\newblock {\em {IEEE} Trans. Robot.}, 37(2):466--481, 2021.

\bibitem{Yu2022ATW}
Zheng Yu and et~al.
\newblock A tapered whisker-based physical reservoir computing system for mobile robot terrain identification in unstructured environments.
\newblock {\em IEEE Robotics and Automation Letters}, 7:3608--3615, 2022.

\bibitem{active}
Gabriel~B Margolis and et~al.
\newblock Learning to see physical properties with active sensing motor policies.
\newblock {\em Conference on Robot Learning}, 2023.

\bibitem{dinov1}
Mathilde Caron and et~al.
\newblock Emerging properties in self-supervised vision transformers.
\newblock In {\em Proceedings of the International Conference on Computer Vision (ICCV)}, 2021.

\bibitem{dinov2}
Maxime Oquab and et~al.
\newblock Dinov2: Learning robust visual features without supervision, 2023.

\bibitem{Anqiao23}
Anqiao Li and et~al.
\newblock Seeing through the grass: Semantic pointcloud filter for support surface learning.
\newblock {\em IEEE Robotics and Automation Letters}, 8(11):7687--7694, 2023.

\bibitem{richter}
Charles Richter and Nicholas Roy.
\newblock Safe visual navigation via deep learning and novelty detection.

\bibitem{slic}
Radhakrishna Achanta and et~al.
\newblock Slic superpixels compared to state-of-the-art superpixel methods.
\newblock {\em IEEE Transactions on Pattern Analysis and Machine Intelligence}, 34(11):2274--2282, 2012.

\bibitem{GMM}
Douglas Reynolds.
\newblock {\em Gaussian Mixture Models}, pages 659--663.
\newblock Springer US, Boston, MA, 2009.

\bibitem{kingma2014adam}
Diederik~P Kingma and Jimmy Ba.
\newblock Adam: A method for stochastic optimization.
\newblock {\em arXiv preprint arXiv:1412.6980}, 2014.

\bibitem{bloesch2013state}
Michael Bloesch and et~al.
\newblock State estimation for legged robots: Consistent fusion of leg kinematics and imu.
\newblock 2013.

\bibitem{Haddeler2022RealtimeDD}
Garen Haddeler and et~al.
\newblock Real-time digital double framework to predict collapsible terrains for legged robots.
\newblock {\em 2022 IEEE/RSJ International Conference on Intelligent Robots and Systems (IROS)}, pages 10387--10394, 2022.

\end{thebibliography}

% You can push biographies down or up by placing
% a \vfill before or after them. The appropriate
% use of \vfill depends on what kind of text is
% on the last page and whether or not the columns
% are being equalized.

%\vfill

% Can be used to pull up biographies so that the bottom of the last one
% is flush with the other column.
%\enlargethispage{-5in}

% that's all folks
\end{document}